\documentclass{eptcs}
\usepackage{mathptmx}
\usepackage{enumitem}
\setlist[itemize]{noitemsep, topsep=0pt}
\usepackage[utf8]{inputenc}
\usepackage{graphicx}
\usepackage[final]{listings}
\usepackage{float}
\usepackage{tikz}
\usetikzlibrary{calc}
\usepackage{url}
\usepackage{multirow}
\usepackage{multicol}
\usepackage{booktabs}

\usepackage[final,commentmarkup=todo,authormarkup=brackets,highlightmarkup=uwave]{changes}

\definechangesauthor{FC}
\definechangesauthor{MRV}

\usepackage{subcaption}
\usepackage{amsmath,amssymb}
\usepackage{ulem}

\usepackage{todonotes}
\presetkeys
    {todonotes}
    {inline,backgroundcolor=gray}{}

\newcommand\pureNN{Pure NN}

\newfloat{lstfloat}{htbp}{lop}
\floatname{lstfloat}{Listing}

\usetikzlibrary{positioning,calc,fit}

\author
    {Marc Roig Vilamala, Harrison Taylor
    \institute{Cardiff University}
    \and Tianwei Xing, Luis Garcia, Mani Srivastava
    \institute{University of California, Los Angeles}
    \and Lance Kaplan
    \institute{CCDC Army Research Laboratory}
    \and Alun Preece, Angelika Kimmig
    \institute{Cardiff University}
    \and Federico Cerutti
    \institute{University of Brescia}}

\title{A Hybrid Neuro-Symbolic Approach for Complex Event Processing\thanks{This research was sponsored by the U.S. Army Research Laboratory and the U.K. Ministry of Defence under Agreement Number W911NF-16-3-0001. The views and conclusions contained in this document are those of the authors and should not be interpreted as representing the official policies, either expressed or implied, of the U.S. Army Research Laboratory, the U.S. Government, the U.K. Ministry of Defence or the U.K. Government. The U.S. and U.K. Governments are authorized to reproduce and distribute reprints for Government purposes notwithstanding any copyright notation hereon.}}
\def\titlerunning{A Hybrid Neuro-Symbolic Approach for Complex Event Processing}
\def\authorrunning{Marc Roig Vilamala et al.}

\begin{document}

\maketitle

\begin{abstract}
    Training a model to detect patterns of interrelated events that form situations of interest can be a complex problem: such situations tend to be uncommon, and only sparse data is available.
    We propose a hybrid neuro-symbolic architecture based on Event Calculus that can perform Complex Event Processing (CEP). It leverages both a neural network to interpret inputs and logical rules that express the pattern of the complex event.
    Our approach is capable of training with much fewer labelled data than a pure neural network approach, and to learn to classify individual events even when training in an end-to-end manner.
    We demonstrate this comparing our approach against a pure neural network approach on a dataset based on Urban Sounds 8K.
\end{abstract}

\section{Introduction}
\label{sec:introduction}

Imagine a scenario where we are trying to detect a \textit{shooting} using microphones deployed in a city: \textit{shooting} is a situation of interest that we want to identify from a high-throughput (audio) data stream. Complex Event Processing (CEP) is a type of approach aimed at detecting such situations of interest, called \textit{complex events}, from a data stream using a set of rules. These rules are defined on atomic pieces of information from the data stream, which we call events---or \textit{simple events}, for clarity. Complex events can be formed from multiple simple events. For instance, \textit{shooting} might start when multiple instances of the simple event \textit{gunshot} occur.
For simplicity, we can assume that when we start to detect \textit{siren} events,  authorities have arrived and the situation is being dealt with, which would conclude the complex event.

Using the raw data stream implies that usually we cannot directly write declarative rules on that data, as it would imply that we need to process that raw data using symbolic rules;
though theoretically possible, this is hardly recommended.

Using a machine learning algorithm such a neural network trained with back-propagation is also infeasible, as it will need to simultaneously learn to understand the simple events within the data stream, and the interrelationship between such events to compose a complex event. While possible, the sparsity of data makes this a hard problem to solve.

The architecture we propose\footnote{Code is available at \url{https://github.com/MarcRoigVilamala/DeepProbCEP}} is a hybrid neuro-symbolic approach that allows us to combine the advantages of both approaches. Our approach is capable of performing CEP on raw data after training in an end-to-end manner.
Among other advantages, our approach is better at training with sparse data than pure neural network approaches, as we will demonstrate.

\section{Background}
\label{sec:background}

ProbEC \cite{Skarlatidis2015} is an approach for complex event processing using probabilistic logic programming.
ProbEC takes an input stream of simple events---each of which has a probability of happening attached.
From there, it is capable of outputting how likely it is for a complex event to be happening at a given point in time based on some manually-defined rules that describe the pattern for the complex event.

In a previous paper, we built on top of ProbEC
proposing a system that made use of pre-trained neural networks to detect complex events from CCTV feeds \cite{Fusion}.
However this approach required access to pre-trained neural networks to process the simple events, which are not always available. To solve this issue, we moved towards end-to-end training, which is able to train these neural networks from just the input data and labels on when the complex events are happening.

In order to implement an end-to-end training with a hybrid neuro-symbolic approach, we made use of \textbf{DeepProbLog} \cite{deepproblognips}, which incorporates deep learning into a probabilistic programming language. This allowed us to train the neural networks as part of the system in an end-to-end manner.

DeepProbLog allows users to make use of the outputs of a neural network as part of the knowledge database in a ProbLog program. DeepProbLog also allows users to train those neural networks in an end-to-end manner by calculating the gradient required to perform the gradient descent based on the true value of the performed query and the outputs provided by the neural network. This allows us to implement a hybrid neuro-symbolic architecture that is able to learn in an end-to-end manner.

\section{Our Approach}
\label{sec:methodology}

In this paper, we are proposing a hybrid neuro-symbolic architecture that performs CEP.
As a proof of concept, we have implemented our architecture to perform CEP on audio data.
In our implementation, each audio second is processed by a PyTorch implementation of VGGish\footnote{Available at \url{https://github.com/harritaylor/torchvggish}}, a feature embedding frontend for audio classification models \cite{hershey2017cnn} which outputs a feature vector for each second of the original audio file.
We use these features as input of a multi-layer perceptron (MLP, AudioNN in the figure) that classifies the audio into a pre-defined set of classes.

\newcommand{\nndraw}[1]{
\def\layersep{2.5cm}
\begin{tikzpicture}[node distance=\layersep,scale={#1}, every node/.style={scale={#1}}]

    \tikzstyle{neuron}=[circle,minimum size=0.2pt,inner sep=0pt,line width=0.05mm,draw,scale=0.3]
    \tikzstyle{input neuron}=[neuron]
    \tikzstyle{output neuron}=[neuron]
    \tikzstyle{hidden neuron}=[neuron]

    \foreach \name / \y in {1,...,4}

        \node[input neuron] (I-\name) at (0,-\y) {};

    \foreach \name / \y in {1,...,5}
        \path[yshift=0.5cm]
            node[hidden neuron] (H-\name) at (\layersep,-\y cm) {};

    \foreach \name / \y in {1/2,2/3,3/4}
        \path[yshift=0.5cm]
            node[output neuron] (O-\name) at (2*\layersep,-\y cm) {};

    \foreach \source in {1,...,4}
        \foreach \dest in {1,...,5}
            \path (I-\source) edge (H-\dest);

    \foreach \source in {1,...,5}
        \foreach \dest in {1,...,3}
            \path (H-\source) edge (O-\dest);

\end{tikzpicture}}

\newcommand{\mlpdraw}[1]{
\def\layersep{2.5cm}
\begin{tikzpicture}[node distance=\layersep,scale={#1}, every node/.style={scale={#1}}]

    \tikzstyle{neuron}=[circle,minimum size=0.2pt,inner sep=0pt,line width=0.05mm,draw,scale=0.2]
    \tikzstyle{input neuron}=[neuron]
    \tikzstyle{output neuron}=[neuron]
    \tikzstyle{hidden neuron}=[neuron]

    \foreach \name / \y in {1,...,5}

        \node[input neuron] (I-\name) at (0,-\y) {};

    \foreach \name / \y in {1,...,4}
        \path[yshift=-0.5cm]
            node[hidden neuron] (H-\name) at (\layersep,-\y cm) {};

    \foreach \name / \y in {1/2,2/3,3/4}
        \path[yshift=0.0cm]
            node[output neuron] (O-\name) at (2*\layersep,-\y cm) {};

    \foreach \source in {1,...,5}
        \foreach \dest in {1,...,4}
            \path (I-\source) edge (H-\dest);

    \foreach \source in {1,...,4}
        \foreach \dest in {1,...,3}
            \path (H-\source) edge (O-\dest);

\end{tikzpicture}}

The output of this neural network is then used in the logic layer, which contains the rules required to perform CEP. Here, the user can define which patterns of simple events constitute the starts and ends of which complex events.

DeepProbLog is used to integrate the different parts, which allows us to train the neural network within the system in an end-to-end manner, which heavily reduces the cost of labelling; it is practically infeasible to label each second of large collections of audio tracks, while it is much easier to identify the beginning and the end of complex events as situations of interest. As such the system is provided with raw audio data and, for training, labels on when the complex events start and end.

We experimentally compare our approach against
a pure statistical learning approach using a neural network (\pureNN{}).
\pureNN\ exchanges the logical layer for an MLP, which will learn the rules that define complex events.
We engineered a synthetic dataset based on Urban Sounds 8K \cite{UrbanSound}.
We consider two repetitions of the same start event---or end event--- within a certain time window as the signal of the beginning---or termination--- of a fluent. Then, to test the efficacy of the approach, we varied the size of the time window for repetitions, from 2 to 5 seconds.
The information on when a complex event begins or ends is used as training data.
The goal of our sysntetic dataset is to be able to detect when a complex event is happening from raw data.

For all the reported results, the corresponding system has been trained on a sequence generated from randomly-ordering the files from 9 of the 10 pre-sorted folds from Urban Sounds 8K, with the remaining fold being used for testing. As recommended by the creators of the dataset, 10-fold cross validation has been used for evaluation.
Before each evaluation, both systems have been trained for 10 epochs with 750 training points on each epoch. This allows for both approaches to converge according to our experiments.

\begin{table}[]
\centering
\caption{Accuracy results (average over 10-fold cross validation) for a hybrid neuro-symbolic architecture (our approach) and a pure neural network approach for both individual sounds (Sound Acc) and a pattern of two instances of the same sound class (Pattern Acc) within the window size. Best in bold.
}
\label{tbl:audio_results}
\begin{tabular}{cccccc}
\toprule
                                  & \multirow{2}{*}{Approach}       & \multicolumn{4}{c}{Window Size}               \\
                                  &                                 & 2         & 3         & 4         & 5         \\
\midrule
\multirow{2}{*}{Sound Accuracy}   & Hybrid (Our approach) & \textbf{0.6425} & \textbf{0.5957} & \textbf{0.6157} & \textbf{0.6076} \\
                                  & \pureNN{}      & 0.0725          & 0.1155          & 0.0845          & 0.0833          \\
\hline
\multirow{2}{*}{Pattern Accuracy} & Hybrid (Our approach) & \textbf{0.5180} & \textbf{0.4172} & \textbf{0.4624} & \textbf{0.4506} \\
                                  & \pureNN{}      & 0.1843          & 0.2034          & 0.2289          & 0.1927 \\
\bottomrule
\end{tabular}
\end{table}

Table \ref{tbl:audio_results} shows the results of our approach and \pureNN{} with different window sizes. It shows both the performance for detecting the starts and ends of complex events (Pattern Accuracy) and for classifying the simple events in the sequence (Sound Accuracy).
As we can see in the table, our approach is clearly superior, as \pureNN{} has a performance only marginally better than a random classifier, which would archive a performance of about $10\%$.
Therefore
our approach is very efficient at learning from sparse data, and, as a byproduct, can also train the neural network to classify simple events.

\section{Conclusions}
\label{sec:conclusion}

In this paper we demonstrated the superiority of our approach against a feedforward neural architecture. Further investigations that consider recurrent networks (RNNs) particularly long-short term memory (LSTM) networks are ongoing; we thank an anonymous reviewer for this suggestion.
Further research could also include rule learning, which could be used to remove the necessity of knowing which patterns of simple events form which complex events.

\bibliographystyle{eptcs}
\bibliography{biblio}

\begin{thebibliography}{1}
\providecommand{\bibitemdeclare}[2]{}
\providecommand{\surnamestart}{}
\providecommand{\surnameend}{}
\providecommand{\urlprefix}{Available at }
\providecommand{\url}[1]{\texttt{#1}}
\providecommand{\href}[2]{\texttt{#2}}
\providecommand{\urlalt}[2]{\href{#1}{#2}}
\providecommand{\doi}[1]{doi:\urlalt{http://dx.doi.org/#1}{#1}}
\providecommand{\bibinfo}[2]{#2}

\bibitemdeclare{inproceedings}{hershey2017cnn}
\bibitem{hershey2017cnn}
\bibinfo{author}{S.~\surnamestart Hershey\surnameend},
  \bibinfo{author}{S.~\surnamestart Chaudhuri\surnameend},
  \bibinfo{author}{D.~P.~W. \surnamestart Ellis\surnameend},
  \bibinfo{author}{J.~F \surnamestart Gemmeke\surnameend},
  \bibinfo{author}{A.~\surnamestart Jansen\surnameend}, \bibinfo{author}{R.~C.
  \surnamestart Moore\surnameend}, \bibinfo{author}{M.~\surnamestart
  Plakal\surnameend}, \bibinfo{author}{D.~\surnamestart Platt\surnameend},
  \bibinfo{author}{R.~A \surnamestart Saurous\surnameend},
  \bibinfo{author}{B.~\surnamestart Seybold\surnameend} et~al.
  (\bibinfo{year}{2017}): \emph{\bibinfo{title}{CNN architectures for
  large-scale audio classification}}.
\newblock In: {\sl \bibinfo{booktitle}{2017 IEEE International Conference on
  Acoustics, Speech and Signal Processing}}, \bibinfo{organization}{IEEE}, pp.
  \bibinfo{pages}{131--135}.

\bibitemdeclare{incollection}{deepproblognips}
\bibitem{deepproblognips}
\bibinfo{author}{Robin \surnamestart Manhaeve\surnameend},
  \bibinfo{author}{Sebastijan \surnamestart Dumancic\surnameend},
  \bibinfo{author}{Angelika \surnamestart Kimmig\surnameend},
  \bibinfo{author}{Thomas \surnamestart Demeester\surnameend} \&
  \bibinfo{author}{Luc \surnamestart De~Raedt\surnameend}
  (\bibinfo{year}{2018}): \emph{\bibinfo{title}{DeepProbLog: Neural
  Probabilistic Logic Programming}}.
\newblock In: {\sl \bibinfo{booktitle}{NIPS2018}}, pp.
  \bibinfo{pages}{3749--3759}.

\bibitemdeclare{inproceedings}{Fusion}
\bibitem{Fusion}
\bibinfo{author}{M.~\surnamestart {Roig Vilamala}\surnameend},
  \bibinfo{author}{L.~\surnamestart {Hiley}\surnameend},
  \bibinfo{author}{Y.~\surnamestart {Hicks}\surnameend},
  \bibinfo{author}{A.~\surnamestart {Preece}\surnameend} \&
  \bibinfo{author}{F.~\surnamestart {Cerutti}\surnameend}
  (\bibinfo{year}{2019}): \emph{\bibinfo{title}{A Pilot Study on Detecting
  Violence in Videos Fusing Proxy Models}}.
\newblock In: {\sl \bibinfo{booktitle}{2019 22th International Conference on
  Information Fusion}}, pp. \bibinfo{pages}{1--8}.

\bibitemdeclare{inproceedings}{UrbanSound}
\bibitem{UrbanSound}
\bibinfo{author}{J.~\surnamestart Salamon\surnameend},
  \bibinfo{author}{C.~\surnamestart Jacoby\surnameend} \&
  \bibinfo{author}{J.~P. \surnamestart Bello\surnameend}
  (\bibinfo{year}{2014}): \emph{\bibinfo{title}{A Dataset and Taxonomy for
  Urban Sound Research}}.
\newblock In: {\sl \bibinfo{booktitle}{22nd {ACM} International Conference on
  Multimedia (ACM-MM'14)}}, \bibinfo{address}{Orlando, FL, USA}, pp.
  \bibinfo{pages}{1041--1044}.

\bibitemdeclare{article}{Skarlatidis2015}
\bibitem{Skarlatidis2015}
\bibinfo{author}{A.~\surnamestart Skarlatidis\surnameend},
  \bibinfo{author}{A.~\surnamestart Artikis\surnameend},
  \bibinfo{author}{J.~\surnamestart Filippou\surnameend} \&
  \bibinfo{author}{G.~\surnamestart Paliouras\surnameend}
  (\bibinfo{year}{2015}): \emph{\bibinfo{title}{{A probabilistic logic
  programming event calculus}}}.
\newblock {\sl \bibinfo{journal}{Theory and Practice of Logic Programming}}
  \bibinfo{volume}{15}(\bibinfo{number}{2}), pp. \bibinfo{pages}{213--245}.

\end{thebibliography}

\end{document}